%% file: neurips_2026.tex
\documentclass{article}

\PassOptionsToPackage{numbers, sort&compress}{natbib}
\usepackage[preprint]{neurips_2026}

\usepackage[utf8]{inputenc}
\usepackage[T1]{fontenc}
\usepackage{hyperref}
\usepackage{url}
\usepackage{booktabs}
\usepackage{amsfonts}
\usepackage{amsmath,amssymb}
\usepackage{mathtools}
\usepackage{microtype}
\usepackage{xcolor}
\usepackage{graphicx}
\usepackage{colortbl}
\usepackage{multirow}
\usepackage{enumitem}
\usepackage{caption}
\usepackage{float}
\usepackage{placeins}
\definecolor{linkred}{RGB}{170,30,30}
\definecolor{citeblue}{RGB}{30,80,170}
\hypersetup{colorlinks=true, linkcolor=linkred, citecolor=citeblue, urlcolor=citeblue}

\title{GenRM-Flow: Generators are Process-aware Reward Models in Flow Matching}

\author{%
  Siming Fu\textsuperscript{1*} \quad
  Ruizhe He\textsuperscript{1*\dag} \quad
  Zheming Fu\textsuperscript{2*} \quad
  Zeyue Xue\textsuperscript{2} \quad
  Fengkai Liu\textsuperscript{2} \quad
  Yufei Jiao\textsuperscript{3} \\
  \bfseries Jie Huang\textsuperscript{2} \quad
  Xiaoxiao Ma\textsuperscript{2} \quad
  Xuan Yang\textsuperscript{2} \quad
  Lihui Luo\textsuperscript{4} \\[4pt]
  {\normalfont \textsuperscript{1}Zhejiang University \quad \textsuperscript{2}Joy Future Academy \quad \textsuperscript{3}Xiamen University \quad \textsuperscript{4}Tsinghua University} \\
  {\normalfont\texttt{fusiming@zju.edu.cn} \qquad \texttt{22432120@zju.edu.cn}}
}

\begin{document}

\maketitle
{\renewcommand{\thefootnote}{\fnsymbol{footnote}}
\footnotetext[1]{Equal contribution.}
\footnotetext[2]{Corresponding author.}}

\vspace{-0.25in}
\begin{figure}[H]
    \centering
    \includegraphics[width=\linewidth]{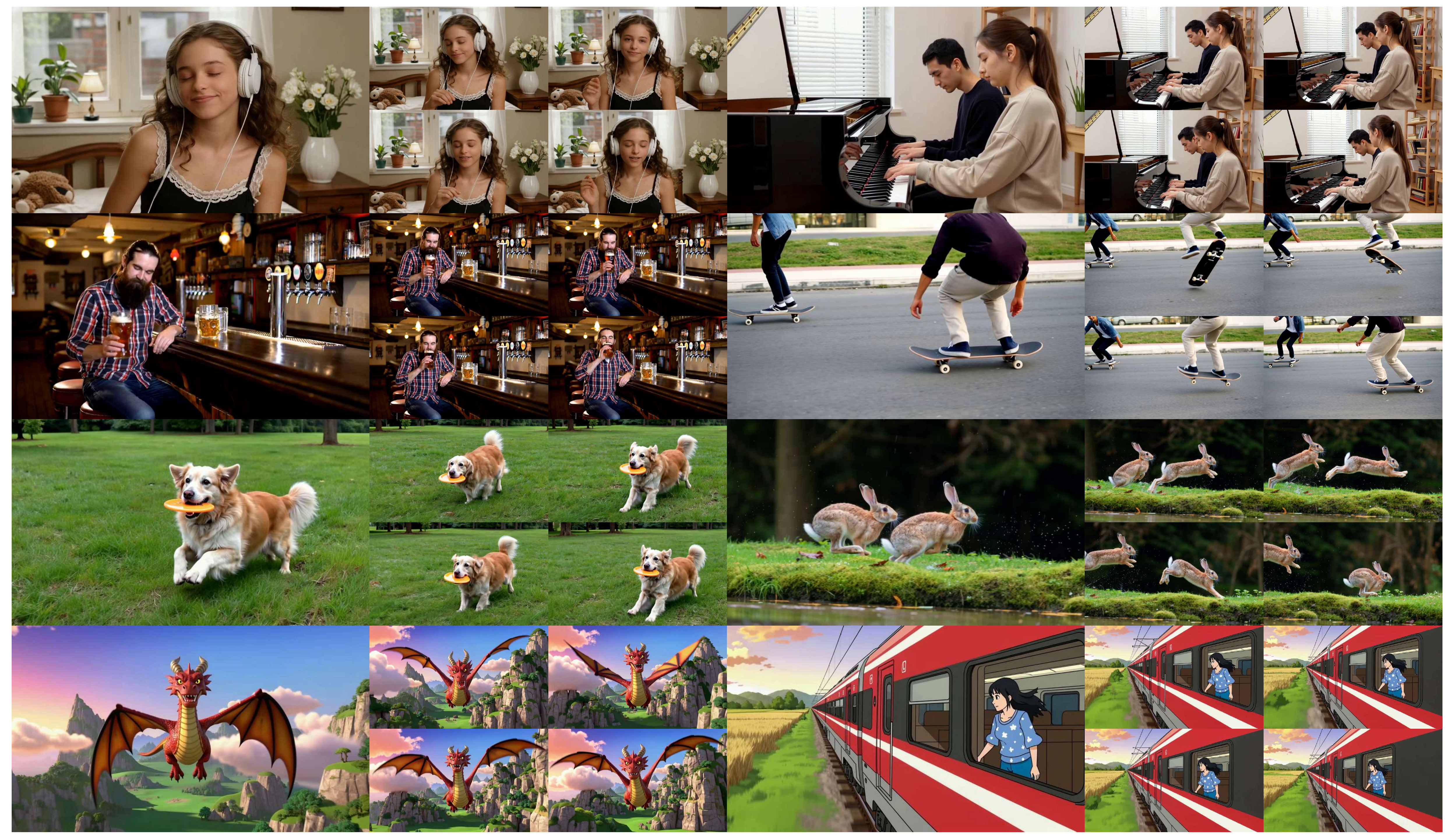}
    \caption{Videos generated by Wan2.1-T2V-14B fine-tuned with FlowAWR using the GenRM-Flow reward, exhibiting high motion quality and dynamic fidelity.}
    \label{fig:teaser}
\end{figure}

\begin{abstract}
Aligning video generation models with human preference depends critically on the quality of the reward model that scores generated videos. Current video reward models, however, source their reward signal from a model's representational rather than generative capacity: they either repurpose vision-language models as pixel-space scorers, or train auxiliary scoring heads on top of frozen generator features. The generator's actual ability to predict velocities, denoise latents, and produce videos plays no role in scoring its own generations. We argue that this overlooks a structural property of offline preference learning. Any preference optimization objective that approximates an offline preference dataset mathematically targets a strictly defined KL-constrained optimal policy. By isolating the reward term from this analytical solution and mapping the exact generation log-likelihood to its flow-matching surrogate, the empirical reward emerges deterministically as the relative difference in continuous-time velocity-prediction errors. We formalize this equivalence into \textbf{GenRM-Flow}, a framework where a preference-tuned video generator inherently operates as a process-aware reward model directly in the noisy latent space. We further show theoretically that distinct fine-tuning objectives, whether contrastive like DPO and IPO or non-contrastive like RWR, are alternative optimization pathways toward the same optimal policy, each inducing a reward readout from the generator's velocity-prediction errors. Furthermore, the extracted process reward seamlessly substitutes external VLM scorers in FlowAWR. Operating purely on native latent outputs without auxiliary reward networks, this generative reward drives measurable improvements in downstream video generation.
\end{abstract}

\input{sec/Introduction}
\input{sec/Related_work}
\input{sec/Method}
\input{sec/Experiment}
\FloatBarrier
\input{sec/Conclusion}

{\small
\bibliographystyle{plainnat}
\bibliography{reference}
}

\end{document}

%% file: sec/Introduction.tex
\section{Introduction}
\label{sec:intro}

Video generation models~\cite{kong2024hunyuanvideo,wan2025wan,gao2025seedance} have reached a level of fidelity where the next bottleneck is no longer raw quality but \emph{alignment}: making the model produce videos that match human preference along axes such as motion smoothness, prompt alignment, and aesthetic quality. As in language models, alignment in video generation is increasingly framed as a reward modeling problem. Preference data is collected, a reward model (RM) is trained to score generated videos, and the generator is updated through preference optimization~\cite{wallace2024diffusion,liu2025improving} or reward feedback learning~\cite{mi2025video,liu2025flow}. The reward model is the load-bearing component of this pipeline: any error in how it scores videos translates directly into degradation of the trained generator.

Despite this central role, the shape of existing video reward models follows from a single root commitment: \emph{the reward is computed by a model other than the generator itself}, and the rest of the architecture is largely determined by this choice.
The VLM-based paradigm~\cite{liu2025improving}, which sources rewards from a model trained on RGB inputs, requires a noisy latent to be decoded into pixels.
VAE decoding fails in high-noise regimes, restricting VLM rewards to late-stage timesteps and leaving early structure and motion formation unsupervised. 
The feature-extractor paradigm~\cite{mi2025video} bypasses pixels but evaluates uncalibrated intermediate features, necessitating a scoring head with independent objectives. 
This places the reward in a disjoint parameter space, where parametric divergence directly degrades alignment. \textbf{\textit{Across both paradigms, the generator's own capacity is decoupled from the evaluation phase, structurally omitting a fundamental source of alignment signal.}}

We close this gap by introducing \textbf{GenRM-Flow}, a reward modeling framework in which the video generator itself serves as the process-aware reward model. 
The starting observation is a structural property of offline preference learning: any preference optimization designed to approximate a preference dataset mathematically targets a strictly defined KL-constrained optimal policy.
By isolating the reward term from this analytical solution and utilizing the flow-matching velocity-prediction error as a surrogate for generation log-likelihood, a unified empirical reward formulation emerges. 
Because these preference objectives structurally target the identical optimal density shift, the implicit reward is consistently extracted directly from the generator's continuous-time denoising errors. 
After training, this evaluation is computed from the same velocity predictions the generator already produces, contrasted against a frozen reference model. Three structural properties fall out of this construction rather than requiring engineering. 
First, it operates natively in the noisy latent space without VAE decoding bottlenecks. Second, it is intrinsically process-aware, as the continuous-time integral is mathematically decomposable across any integration step. 
Most distinctively, the reward signal is sourced purely from the model's \emph{generative capacity}, requiring no auxiliary prediction head or separate evaluation server.

We validate GenRM-Flow in two ways. 
First, the resulting reward model achieves strong discriminative accuracy on held-out preference pairs, outperforming both VLM-based and feature-extractor-based reward models, with a consistent margin over the latter across all noise levels.
Second, the reward model is operationally effective for online policy improvement. Plugged into FlowAWR~\cite{fu2026flowawr} as the per-step process reward source, we demonstrate that by substituting true data with generated end-states as trajectory proxy targets and applying group-wise Softmax normalization, under which the intractable partition function cancels, the offline-derived reward signal successfully drives measurable downstream generation improvement. \textbf{We summarize our contributions as follows:}
\begin{itemize}
\item We propose \textbf{GenRM-Flow}, a reward modeling framework that closes a structural gap in current video reward modeling: where prior approaches score generations without using generation, GenRM-Flow trains a video generator to serve as a process-aware reward model through preference optimization alone, with no auxiliary head.
\item We validate the operational efficacy of the resulting reward model by integrating it as a per-step process reward in FlowAWR. We show that with group-wise reward normalization and trajectory proxy targets, this generation-based reward successfully drives sustained online policy improvement.
\end{itemize}

%% file: sec/Related_work.tex
\section{Related Work}
\label{sec:related_work}

\noindent \textbf{Reward Models for Video Generation.}
Reward modeling is now central to aligning generative models~\cite{xu2025drc, xu2025personalized, xu2025personalized_v2, xu2024diffusion, qiu2025accelerating, qiu2025accelerating_v2, wang2025precise, li2026speed, li2026easier, Kling-3.0, hailuo2025} with human preferences. Existing video reward models are predominantly outcome-based~\cite{shao2024deepseekmath}, scoring fully generated outputs: T2VQA~\cite{kou2024subjective} and VideoScore~\cite{he2024videoscore} regress human ratings, VideoAlign~\cite{liu2025improving} trains on pairwise preferences via the Bradley-Terry loss~\cite{bradley1952rank, rao1967ties}, and others adapt VLMs~\cite{wu2024boosting, yuan2024instructvideo, mougradeo} or add chain-of-thought reasoning for interpretability~\cite{wang2025unified2, wang2024lift, he2025videoscore2, wu2025rewarddance, wang2025unified}. Operating only on near-complete outputs, these models provide no guidance during the early denoising stages that determine fundamental video attributes.

Process reward models~\cite{lightman2024lets, wang2024math, zhang2025lessons} address this gap in the language domain by supervising intermediate steps. For visual generation, LPO~\cite{zhang2025diffusion} pioneered using the diffusion model itself as a noise-aware latent reward for step-level preference optimization in images, and PAVRM~\cite{mi2025video} extends this to video by attaching a separate reward head atop frozen VGM features. PAVRM, however, treats the generator as a feature extractor: the reward comes from the auxiliary head rather than the model's own velocity predictions, leaving the generative capacity unused in scoring. In contrast, we derive a fully implicit process reward from the policy model's prediction error relative to the reference, requiring no auxiliary parameters and aligning with the generation process at every denoising timestep.

\noindent \textbf{Reinforcement Learning for Visual Generation.}
The integration of reinforcement learning (RL) into large language models~\cite{hurst2024gpt, hui2024qwen2, jaech2024openai, GPT-5, gemini-2.5-pro} and multimodal LLMs has significantly advanced reasoning capabilities~\cite{r1_reward, vlmr1, internvl3, llavar1}, driven by the shift from replicating training data to refining reasoning trajectories through reward optimization.
Proximal Policy Optimization (PPO)~\cite{ppo, lu2025adavipaligningmultimodalllms} initiated this paradigm, while Group Relative Policy Optimization (GRPO)~\cite{DeepSeekR1, DeepSeekMath} simplified advantage computation by eliminating the critic network.
GRPO has been successfully applied to visual understanding~\cite{yu2024rlaif, zhou2024aligning, liu2025visual, sun2024aligning, yu2024rlhf, lu2025dama, liu2025diversegrpo, li2026enhancing} and, more recently, to reward modeling for visual generation~\cite{wu2025visualquality, he2025videoscore2, li2025q}.
We provide a per-step process reward derived from the VGM's own predictions, which addresses the sparse reward problem in RL-based video generation fine-tuning by supplying dense supervision at every denoising timestep.

%% file: sec/Method.tex
\section{Method}
\label{sec:method}
\subsection{Preliminaries}
\label{sec:pre}

\paragraph{Rectified flow and flow matching.}
Recent video generation models~\cite{kong2024hunyuanvideo,wan2025wan,gao2025seedance} operate in rectified flow~\cite{lipman2023flow,liu2023flow}, which establishes a continuous transport between data and noise distributions. Given clean data $\mathbf{x}_0 \sim p_{\mathrm{data}}(\cdot\mid c)$ and noise $\mathbf{x}_1\sim \mathcal{N}(0,I)$, the framework defines the interpolation
\begin{equation}
    \mathbf{x}_t = (1-t)\mathbf{x}_0 + t\mathbf{x}_1, \quad t \in [0,1],
    \label{eq:flow_forward}
\end{equation}
and trains a velocity field $\mathbf{v}_\theta(\mathbf{x}_t,c,t)$ with
\begin{equation}
    \mathcal{L}_{\mathrm{FM}}(\theta) = \mathbb{E}_{t,\,\mathbf{x}_0,\,\mathbf{x}_1}\!\left[\|\mathbf{v}_\theta(\mathbf{x}_t, c, t) - \mathbf{v}\|^2\right],
    \qquad \mathbf{v} = \mathbf{x}_1 - \mathbf{x}_0.
    \label{eq:flow_loss}
\end{equation}
We write $\mathcal{L}_\theta(\mathbf{x}_t,c) := \|\mathbf{v}_\theta(\mathbf{x}_t,c,t) - \mathbf{v}\|^2$ for the per-sample velocity error throughout. We will rely on a single property of this loss: it is a Monte Carlo surrogate for the negative log-likelihood of $\mathbf{x}_0$ under model $\theta$~\cite{kingma2021variational,kingma2023understanding}. Up to a positive timestep weighting,
\begin{equation}
-\log p_\theta(\mathbf{x}_0\mid c) \;\lesssim\; \mathbb{E}_{t,\mathbf{x}_1}\!\left[\mathcal{L}_\theta(\mathbf{x}_t,c)\right] + \mathrm{const}.
\label{eq:elbo_flow}
\end{equation}
Eq.~\eqref{eq:elbo_flow} lets us treat $-\mathcal{L}_\theta$ as a stand-in for $\log\pi_\theta$, transporting any quantity defined in terms of policy log-likelihood directly into the flow matching setting.

\subsection{Generation as Reward via Preference Optimization}
\label{subsec:framework}

The starting point is a structural observation about offline preference optimization. Given a static preference dataset $\mathcal{D}^D = \{(c, \mathbf{x}_w, \mathbf{x}_l)\}$ ranked along a target dimension $D$, a broad spectrum of fine-tuning objectives fundamentally optimize the generator to approximate a strictly defined optimal offline policy. These objectives include both contrastive methods, such as DPO~\cite{rafailov2023direct} and IPO~\cite{azar2024general}, and non-contrastive approaches like reward-weighted regression (RWR)~\cite{liu2025improving}. Regardless of their specific loss formulations, these methods implicitly or explicitly target the analytical solution of a KL-constrained reward maximization problem (following the standard derivation in~\cite{rafailov2023direct}):
\begin{equation}
    \pi^*(\mathbf{x}|c) = \frac{1}{Z(c)} \pi_{\mathrm{ref}}(\mathbf{x}|c) \exp\!\left(\frac{R^D(\mathbf{x},c)}{\beta}\right),
    \label{eq:optimal_policy}
\end{equation}
where $Z(c)$ is the partition function and $\beta$ controls the KL-divergence penalty. Because the optimal policy $\pi^*$ is mathematically defined by scaling the reference distribution $\pi_{\mathrm{ref}}$ via the exponentiated reward, updating the policy parameters $\theta$ to approximate $\pi^*$ inherently forces the generator to internalize the preference signal. Consequently, an implicit scalar function $\hat{r}_{\mathcal{F}}(\mathbf{x},c;\theta,\theta_{\mathrm{ref}})$ emerges directly from the density shift dictated by the chosen objective $\mathcal{F}$. Minimizing the respective loss is mathematically equivalent to calibrating this implicit function on the static dataset: the optimization provably fits $\theta$ such that $\hat{r}_{\mathcal{F}}$ evaluates $\mathbf{x}_w$ systematically higher than $\mathbf{x}_l$.

\begin{figure}[t]
  \centering
  \includegraphics[width=\textwidth]{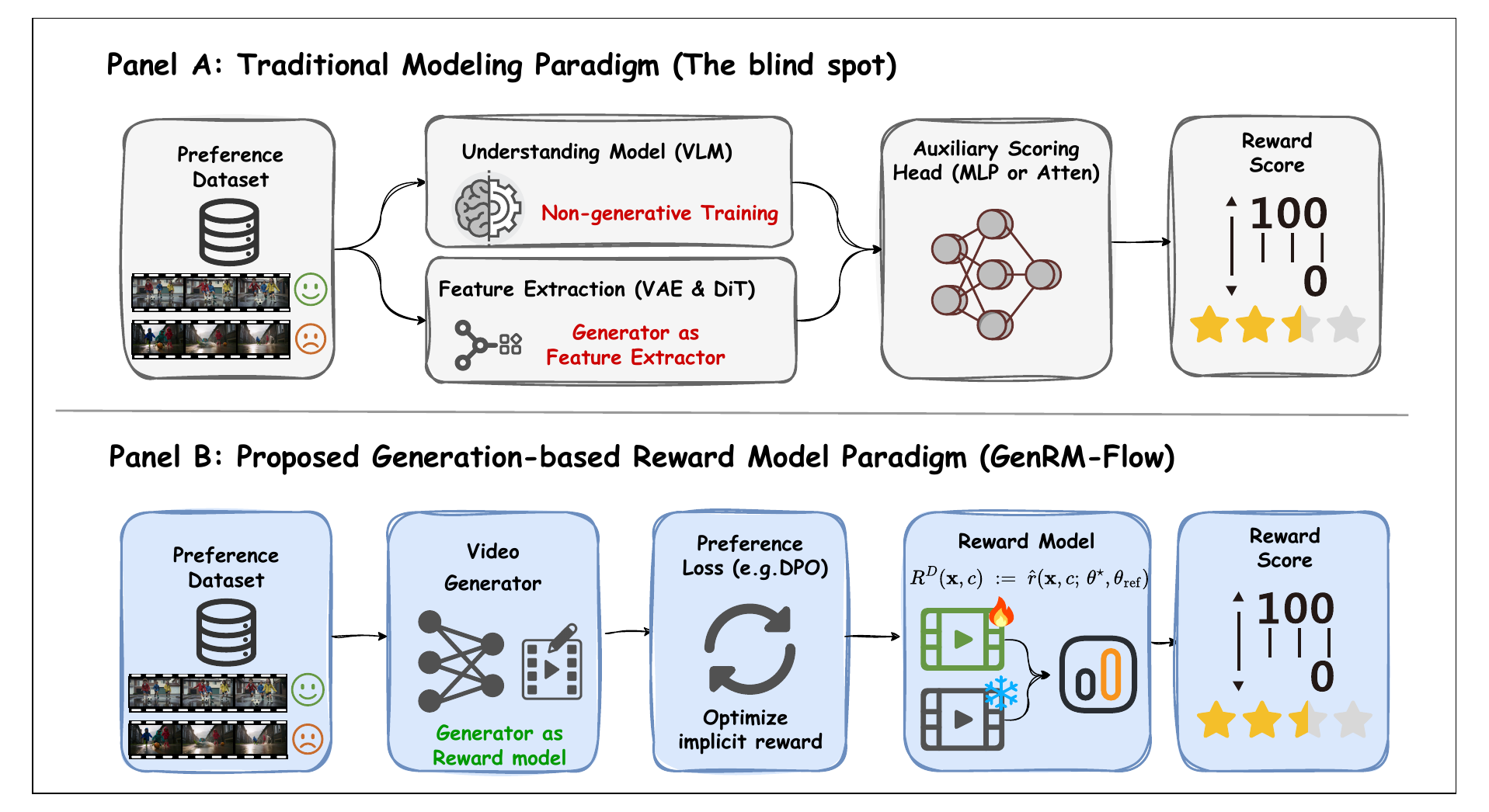}
  \vspace{-6mm}
  \caption{The pipeline of \textbf{GenRM-Flow}. \textbf{(Panel A)} Traditional paradigms rely on \emph{representational capacity}, utilizing a frozen backbone (e.g., a VLM or base generator) as a feature extractor coupled with an auxiliary discriminative head. \textbf{(Panel B)} The proposed framework sources the reward directly from \emph{generative capacity}. Following preference optimization, the implicit reward $R^D(\mathbf{x},c)$ is algebraically extracted from the velocity-prediction errors. The scalar score is computed natively by contrasting the active policy (flame, $\theta^\star$) against a frozen reference (snowflake, $\theta_{\mathrm{ref}}$), requiring no architectural modifications or auxiliary heads.}
  \vspace{-4pt}
  \label{fig:pipeline}
\end{figure}

Consequently, preference optimization achieves a dual objective without architectural modification. Explicitly, it aligns the generative distribution to favor $D$-preferred outputs during sampling. This alignment process structurally configures the scalar function $\hat{r}_{\mathcal{F}}$ to numerically rank inputs along dimension $D$, which aligns with the function of a reward model. Upon the completion of training, this scalar function is implicitly contained within the updated weights, accessible via a standard forward pass, which forms our central claim: \textit{\textbf{A video generation model trained on preference data inherently functions as a reward model.}} Figure~\ref{fig:pipeline} contrasts this generation-based paradigm with existing representation-based reward models.

Three structural properties substantiate this claim. First, the reward derives directly from \emph{generative capacity} rather than intermediate feature extraction, for $\hat{r}_{\mathcal{F}}$ is a strictly defined function of the generation likelihood (or the velocity-prediction error $\mathcal{L}_\theta$ via Eq.~\eqref{eq:elbo_flow}). Second, the framework requires no architectural modification or auxiliary prediction heads. The trained policy $\theta^\star$, whether fully fine-tuned or adapted via LoRA, maintains exact topological equivalence with the base generator, computing scalar evaluations algebraically from native tensor outputs. Third, it eliminates the infrastructure overhead of independent reward modeling. By repurposing the generative computational graph, generation and evaluation are unified within the same system instance, requiring no separate reward server or standalone discriminator network.

\subsection{Empirical Reward via Velocity-Prediction Discrepancy}
\label{subsec:instantiation}

We formalize this structural equivalence algebraically by instantiating the theoretical reward within the flow matching framework. By isolating the reward term from Eq.~\eqref{eq:optimal_policy}, the theoretical reward function is mathematically guaranteed to be:
\begin{equation}
    R^D(\mathbf{x},c) = \beta \log \frac{\pi^*(\mathbf{x}|c)}{\pi_{\mathrm{ref}}(\mathbf{x}|c)} + \beta \log Z(c).
    \label{eq:reward_derivation}
\end{equation}
The partition function $Z(c)$ introduces an intractable, prompt-specific baseline offset. Because $Z(c)$ is strictly dependent on the condition $c$ and invariant to $\mathbf{x}$, it is  neutralized in downstream applications (e.g., via relative ranking or explicit baseline subtraction, as detailed in Section~\ref{subsec:rl_validation}). Consequently, the operative component of the reward is isolated strictly to the log-likelihood ratio.

Assuming the preference optimization successfully drives the parameterized policy $\pi_{\theta^\star}$ to approximate $\pi^*$, we approximate the exact generation log-likelihood using its tractable flow-matching surrogate via the inequality established in Eq.~\eqref{eq:elbo_flow}. By treating $-\mathcal{L}_\theta$ as a bounded proxy for $\log\pi_\theta$, an empirical reward emerges as the expected difference in velocity-prediction errors. Note that this formulation is an approximation subject to the tightness of the variational bound:
\begin{equation}
\hat{r}(\mathbf{x},c;\theta^\star,\theta_{\mathrm{ref}}) \;\approx\; \beta\,\mathbb{E}_{t,\,\mathbf{x}_1}\!\left[\mathcal{L}_{\mathrm{ref}}(\mathbf{x}_t,c) - \mathcal{L}_{\theta^\star}(\mathbf{x}_t,c)\right] + \beta \log Z(c).
\label{eq:rm_empirical}
\end{equation}
It exposes the operational mechanism: a sample receives a higher scalar evaluation if the trained model $\theta^\star$ exhibits a lower denoising error than the reference model $\theta_{\mathrm{ref}}$, conditioned on the assumption that the error divergence proportionally reflects the preference density shift.

\paragraph{Generality Across Preference Objectives.}
This theoretical derivation establishes that the reward evaluation is structurally defined by the underlying fine-tuning objective $\mathcal{F}$. Specific preference optimization algorithms yield specific evaluations corresponding to their density-shift mechanisms (summarized in Table~\ref{tab:loss_comparison}). Contrastive objectives, such as DPO~\cite{rafailov2023direct} and IPO~\cite{azar2024general}, inherently train $\theta^\star$ against a base distribution, yielding the relative evaluation paradigm defined in Eq.~\eqref{eq:rm_empirical}. Alternatively, non-contrastive objectives like RWR~\cite{liu2025improving} operate via preferred-only likelihood maximization, bypassing the reference policy entirely to yield an absolute (reference-free) reward evaluation of the form $R^D(\mathbf{x},c) \propto -\mathbb{E}_{t,\mathbf{x}_1}[\mathcal{L}_{\theta^\star}(\mathbf{x}_t,c)]$; in practice, we instantiate RWR with binary weights (one for preferred samples and zero otherwise), which reduces the weighted objective to likelihood maximization on preferred samples. Both pathways follow from Eq.~\eqref{eq:optimal_policy}: RWR importance-samples the flow-matching objective under $\pi^*$ with $\pi_{\mathrm{ref}}$ as the proposal, weighting samples by $\exp(R^D/\beta)$, while DPO substitutes Eq.~\eqref{eq:reward_derivation} into the Bradley--Terry model, where $Z(c)$ cancels. Regardless of the specific $\mathcal{F}$, the evaluation strictly relies on the respective generative objective, yielding a unified operational pathway:
\begin{equation}
\underbrace{\mathcal{F} \text{ on } \mathcal{D}^D}_{\text{preference optimization}} \;\;\xrightarrow{\text{minimize}}\;\; \underbrace{\theta^\star = \arg\min_\theta \mathcal{L}_{\mathcal{F}}(\theta)}_{\text{aligned generator}} \;\;\xRightarrow{\text{induces}}\;\; \underbrace{R^D(\mathbf{x},c) := \hat{r}(\mathbf{x},c; \theta^\star, \theta_{\text{ref}})}_{\text{implicit reward model}}.
\label{eq:framework_recipe}
\end{equation}

\paragraph{Paradigm Differentiation.}
The mathematical formulation in Eq.~\eqref{eq:rm_empirical} structurally distinguishes the resulting evaluation framework from existing VLM-based and feature-extractor-based paradigms through two operational properties. First, the reward evaluation is executed natively within the noisy latent space. Because the velocity-prediction error $\mathcal{L}_{\theta^\star}$ operates directly on the intermediate latent state $\mathbf{x}_t$, it systematically bypasses RGB pixel-space transformations. This eliminates the severe computational bottleneck of VAE decoding that constrains standard VLM scorers. Second, the evaluation is intrinsically process-aware. Because the expectation in Eq.~\eqref{eq:rm_empirical} is formulated over continuous timesteps $t \in [0,1]$, the integral is mathematically decomposable. Consequently, a single set of weights can output a calibrated evaluation score at any arbitrary point along the denoising trajectory, including the early high-noise regimes where macroscopic structure and motion dynamics form. While recent feature-extractor methods~\cite{mi2025video} must painstakingly engineer auxiliary prediction heads to approximate these capabilities, the proposed framework inherits them as a direct analytical consequence of the generative objective.

\begin{table}[t]
\centering
\small
\renewcommand{\arraystretch}{1.5}
\caption{Three preference fine-tuning objectives as optimization pathways to obtain the trained weights $\theta^\star$. While their loss formulations $\mathcal{L}_{\mathcal{F}}(\theta)$ vary significantly---ranging from bounded contrastive margins to reference-free likelihood maximization---they all structurally aim to approximate the optimal offline policy $\pi^*$. Consequently, during the evaluation phase, they implicitly define reward readouts corresponding to their theoretical likelihood shifts: the relative readout of Eq.~\eqref{eq:rm_empirical} for contrastive objectives and a reference-free readout for RWR.}
\resizebox{\textwidth}{!}{
\begin{tabular}{l|c|c}
\toprule
Fine-tuning Objective $\mathcal{F}$ & Family & Flow Matching Loss Formulation $\mathcal{L}_{\mathcal{F}}(\theta)$ \\
\midrule
RWR~\cite{liu2025improving} & Preferred-only likelihood & $\mathbb{E}_{t,\mathbf{x}_1}\!\left[\mathcal{L}_\theta(\mathbf{x}_{w,t})\right]$ \\
IPO~\cite{azar2024general} & Identity-margin contrastive & $\mathbb{E}_{t,\mathbf{x}_1}\!\left[\left(\bigl[\bigl(\mathcal{L}_{\mathrm{ref}} - \mathcal{L}_\theta\bigr)\!\bigr|_{\mathbf{x}_{w,t}} - \bigl(\mathcal{L}_{\mathrm{ref}} - \mathcal{L}_\theta\bigr)\!\bigr|_{\mathbf{x}_{l,t}}\bigr] - \frac{1}{2\tau}\right)^2\right]$ \\
DPO~\cite{rafailov2023direct} & BT contrastive (saturating) & $-\mathbb{E}_{t,\mathbf{x}_1}\!\left[\log \sigma\!\left(\beta \!\left[\bigl(\mathcal{L}_{\mathrm{ref}} - \mathcal{L}_\theta\bigr)\!\bigr|_{\mathbf{x}_{w,t}} - \bigl(\mathcal{L}_{\mathrm{ref}} - \mathcal{L}_\theta\bigr)\!\bigr|_{\mathbf{x}_{l,t}}\right]\right)\right]$ \\
\bottomrule
\end{tabular}
}
\label{tab:loss_comparison}
\vspace{-6pt}
\end{table}

\subsection{Process Reward Feedback Learning via FlowAWR}
\label{subsec:rl_validation}

To verify the operational validity of the derived latent-space reward, we integrate it into FlowAWR~\cite{fu2026flowawr}. While standard RLHF paradigms compute a global advantage using external VLM scorers after full VAE decoding, our framework bypasses this bottleneck by directly extracting a process-aware latent advantage $A^t(\mathbf{x}_t|c)$.

During rollout, the sampling policy $\theta_{\mathrm{old}}$ generates $G$ trajectories conditioned on prompt $c$ with a deterministic ODE solver. To compute the intermediate reward, we adopt the policy's final generated latent $\hat{\mathbf{x}}_{0}^{(i)}$ as the proxy ground-truth target for its respective trajectory, and obtain $\mathbf{x}_t^{(i)}$ by forward-noising $\hat{\mathbf{x}}_{0}^{(i)}$ via Eq.~\eqref{eq:flow_forward} at the sampled timesteps. Crucially, the evaluation is performed by the offline-trained reward model $\theta^\star$ (e.g., the frozen DPO LoRA) and the reference $\theta_{\mathrm{ref}}$, which are decoupled from the active generation process. The raw process reward at timestep $t$ is evaluated:
\begin{equation}
r_t(\mathbf{x}_t^{(i)},c) \;=\; \beta\!\left[\mathcal{L}_{\mathrm{ref}}(\mathbf{x}_t^{(i)}, \hat{\mathbf{x}}_{0}^{(i)}, c) - \mathcal{L}_{\theta^\star}(\mathbf{x}_t^{(i)}, \hat{\mathbf{x}}_{0}^{(i)}, c)\right].
\label{eq:process_reward}
\end{equation}
Because $\theta^\star$ acts strictly as an independent evaluator during the RL phase, substituting the true data with the generated end-state $\hat{\mathbf{x}}_{0}^{(i)}$ provides a valid velocity target to measure the trajectory's alignment quality, without introducing self-evaluation bias. Furthermore, Eq.~\eqref{eq:process_reward} represents only the log-likelihood ratio and omits the intractable partition function $\beta \log Z(c)$. Since this term is shared by all trajectories sampled for the same prompt $c$, it cancels under any group-relative normalization. We adopt a Softmax transformation over the group:
\begin{equation}
\tilde{r}_t(\mathbf{x}_t^{(i)},c) = \frac{\exp\!\left(r_t(\mathbf{x}_t^{(i)},c)\right)}{\sum_{j=1}^G \exp\!\left(r_t(\mathbf{x}_t^{(j)},c)\right)}.
\label{eq:softmax_reward}
\end{equation}
Beyond removing the prompt-dependent offset, this transformation bounds the reward within $[0,1]$ and, through its exponential form, emphasizes the top-ranked trajectories in each group, converting the reward signal into relative rankings in the spirit of recent preference-based paradigms such as Pref-GRPO~\cite{wang2025pref}.

The timestep-specific advantage $A^t(\mathbf{x}_t^{(i)}|c) = G\,\tilde{r}_t(\mathbf{x}_t^{(i)},c) - 1$ is subsequently computed from these Softmax scores, following the group-based advantage estimate of FlowAWR. The policy is then updated by regressing $\mathbf{v}_\theta(\mathbf{x}_t^{(i)},c,t)$ toward the advantage-weighted rectified target $\mathbf{v}_{\theta_{\mathrm{old}}} + A^t(\mathbf{x}_t^{(i)}|c)\,(\mathbf{v}^{(i)} - \mathbf{v}_{\theta_{\mathrm{old}}})$, where $\mathbf{v}_{\theta_{\mathrm{old}}} = \mathbf{v}_{\theta_{\mathrm{old}}}(\mathbf{x}_t^{(i)},c,t)$ and $\mathbf{v}^{(i)} = \mathbf{x}_1^{(i)} - \hat{\mathbf{x}}_0^{(i)}$. Trajectory sampling and the remaining optimization settings strictly follow FlowAWR defaults, isolating the efficacy of the offline-derived process reward in driving online optimization.

%% file: sec/Experiment.tex
\section{Experiments}
\label{sec:exp}

\subsection{Experimental Setup}

\paragraph{Dataset.}
We collect \textbf{7,170} text prompts from the Internet, spanning multiple semantic categories including \emph{people, animals, scenery, sports, objects, architecture, food}, and \emph{vehicles}.
To construct preference data, we employ 10 text-to-video generation models (Kling-3.0~\cite{Kling-3.0}, Kling-1.6~\cite{Kling-1.6}, Seedance-1.0~\cite{gao2025seedance}, Sora~\cite{OpenAIsora}, Pika-2.2~\cite{Pika2.2}, StepVideo-T2V~\cite{ma2025stepvideot2vtechnicalreportpractice}, Wan2.2 and Wan2.1~\cite{wan2025wan}, Ruyi~\cite{createai2024ruyi}, and LTX-Video~\cite{HaCohen2024LTXVideo}); for each prompt, 6 to 8 models are randomly selected to generate videos, producing \textbf{48,757} videos in total. All generated videos are resized to a unified resolution of $832 \times 480$ at 16\,fps with 5\,s duration.
40 trained annotators then view pairs of videos generated from the same prompt and record pairwise preferences based on overall motion quality. The annotation protocol evaluates three dimensions: (i)~anatomical correctness, covering the proportional integrity of human and animal bodies and deformation artifacts such as twisted, fused, or missing limbs; (ii)~spatial plausibility, covering contact, structural integrity, penetration, and trajectory consistency of interacting objects; and (iii)~temporal coherence, covering the stability of subject attributes, existence, and morphology across frames as well as the continuity of actions. Pairs judged as ties are discarded, and labels are aggregated by majority voting among multiple annotators.
After filtering low-agreement annotations and rebalancing across prompt categories and model pairs, we obtain \textbf{84,597} preference pairs. We randomly hold out \textbf{4,800} pairs for evaluation; the remaining \textbf{79,797} pairs form the training set $\mathcal{D}$.
All methods are trained and evaluated on the same splits to ensure a fair comparison.

\paragraph{Training Details and Baselines.}
We employ the Wan2.1-T2V family~\cite{wan2025wan} as our foundation models, validating our method on both the 1.3B and 14B variants. For our method, we use the AdamW optimizer~\cite{loshchilov2017decoupled} with a learning rate of $1 \times 10^{-5}$ and LoRA rank 64. The DPO coefficient $\beta$ (Table~\ref{tab:loss_comparison}) is set to $2000$. During FlowAWR training, we set $\beta = 1$ in the process reward of Eq.~\eqref{eq:process_reward}; since this reward enters the Softmax in Eq.~\eqref{eq:softmax_reward}, $\beta$ effectively acts as its inverse temperature. Meanwhile, we benchmark our method against two baselines: VideoAlign~\cite{liu2025improving} and PAVRM~\cite{mi2025video}. For VideoAlign~\cite{liu2025improving}, we use Qwen3-VL-8B as the base model. All videos are sampled at 16\,fps with a full resolution of $832 \times 480$ pixels. For PAVRM~\cite{mi2025video}, the original paper trains only on the 14B backbone; we extend it to both 1.3B and 14B using the same architecture: the first 8 DiT blocks serve as the spatiotemporal feature extractor, followed by a query-based attention aggregation layer and a three-layer MLP. The feature dimension is 5120 for the 14B model and 1536 for 1.3B.

\paragraph{Evaluation.}
We evaluate from two perspectives. For reward model evaluation, we measure pairwise preference prediction accuracy on the held-out 4,800 pairs, reported per timestep bin. To further validate the effectiveness of the reward model, we adopt each reward model as the reward function to train Wan2.1-T2V using FlowAWR~\cite{fu2026flowawr}, and assess the trained models with two protocols:
(1)~\textbf{Generation metrics}: following VBench~\cite{huang2023vbench}, we report motion smoothness (MS), dynamic degree (DD), subject consistency (SC), and human action (HA) on both our inner test set and the out-of-domain VBench prompt suite.
(2)~\textbf{HPSv3}~\cite{ma2025hpsv3}: we track HPSv3 as an external visual-quality metric throughout FlowAWR training. Following~\cite{team2025longcat}, we score all sampled frames and report the average of the top 30\% scoring frames, which alleviates the effect of low scores induced by content inconsistency across temporal transitions.

\subsection{Main Results}
\label{subsec:main_results}

\begin{figure}[t]
\centering
\includegraphics[width=0.32\linewidth]{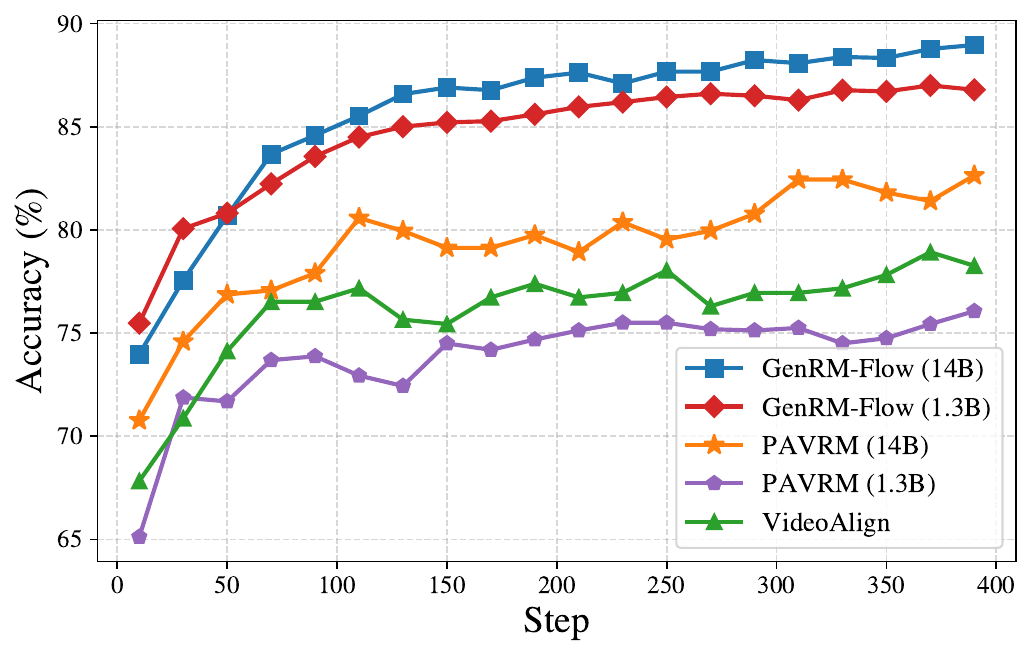}\hfill
\includegraphics[width=0.32\linewidth]{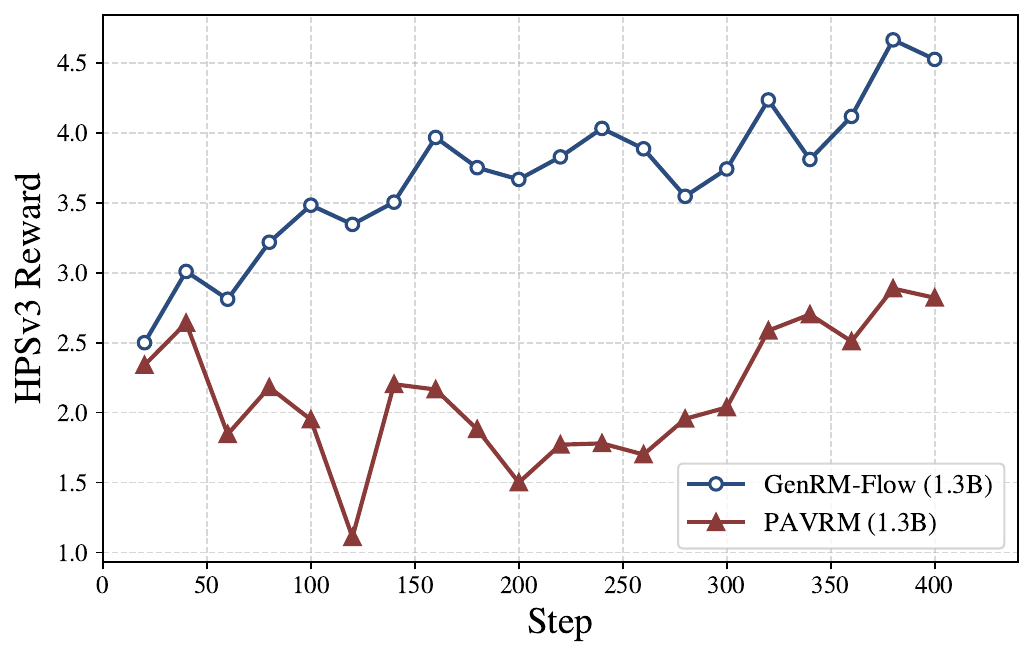}\hfill
\includegraphics[width=0.32\linewidth]{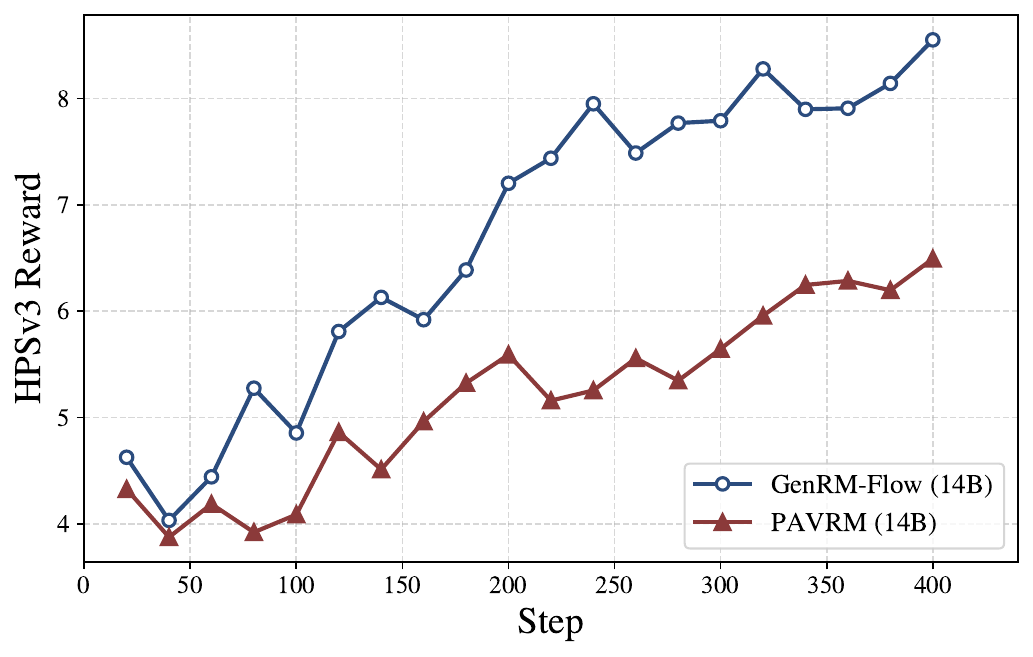}
\caption{(Left) Pairwise preference prediction accuracy during training. (Middle, Right) HPSv3 reward trajectories during FlowAWR training on the 1.3B and 14B backbones. The implicit process reward yields higher sustained HPSv3 evaluation scores compared to PAVRM.}
\label{fig:combined_reward}
\end{figure}

\begin{table*}[!t]
\centering
\caption{Pairwise preference prediction accuracy (\%) on the held-out evaluation set. The performance of our implicit process reward is compared against baseline reward models (VideoAlign and PAVRM) across five uniformly spaced noise timestep intervals $t \in [0, 1]$. \textbf{Bold} denotes the highest accuracy per column within each backbone scale.}
\label{tab:rm_comparison}
\small
\setlength{\tabcolsep}{5.8pt}
\begin{tabular}{llcccccc}
\toprule
\textbf{Base Model} & \textbf{Method} & \textbf{[0, 0.2]} & \textbf{(0.2, 0.4]} & \textbf{(0.4, 0.6]} & \textbf{(0.6, 0.8]} & \textbf{(0.8, 1.0]} & \textbf{Avg} \\
\midrule
Qwen3-VL-8B & VideoAlign~\cite{liu2025improving} & -- & -- & -- & -- & -- & 78.26 \\
\midrule
\multirow{2}{*}{Wan2.1-T2V-1.3B} & PAVRM~\cite{mi2025video} & 77.19 & 77.50 & 78.27 & 76.11 & 71.25 & 76.06 \\
& \textbf{GenRM-Flow} & \textbf{87.77} & \textbf{88.08} & \textbf{88.49} & \textbf{88.61} & \textbf{81.00} & \textbf{86.79} \\
\midrule
\multirow{2}{*}{Wan2.1-T2V-14B} & PAVRM~\cite{mi2025video} & 81.87 & 83.85 & 84.79 & 84.69 & 78.02 & 82.64 \\
& \textbf{GenRM-Flow} & \textbf{89.10} & \textbf{89.94} & \textbf{90.46} & \textbf{90.75} & \textbf{84.55} & \textbf{88.96} \\
\bottomrule
\end{tabular}
\end{table*}

\paragraph{Pairwise Preference Accuracy.}
We compare the pairwise preference prediction accuracy of our implicit reward formulation against PAVRM~\cite{mi2025video} and VideoAlign~\cite{liu2025improving}. As shown in Table~\ref{tab:rm_comparison}, on the 14B backbone, our method reaches $88.96\%$ average accuracy, exceeding PAVRM ($82.64\%$) by $6.32$ points; on the 1.3B backbone, it reaches $86.79\%$ versus $76.06\%$, a $10.73$-point gap. Both configurations also surpass VideoAlign ($78.26\%$). The advantage persists across all five uniformly spaced timestep bins over $t \in [0, 1]$, and remains pronounced in the high-noise regime, where our method sustains at least $81.00\%$ while PAVRM drops to $78.02\%$ (14B) and $71.25\%$ (1.3B).
Figure~\ref{fig:combined_reward} (left) tracks preference accuracy during training: our method converges faster and to a higher terminal value than all baselines on both scales. This suggests that the relative prediction error between policy and reference models provides a stronger preference signal than PAVRM's independent feature-extractor design, while introducing no auxiliary reward-model parameters.

\begin{figure*}[!t]
\centering
\includegraphics[width=1\linewidth]{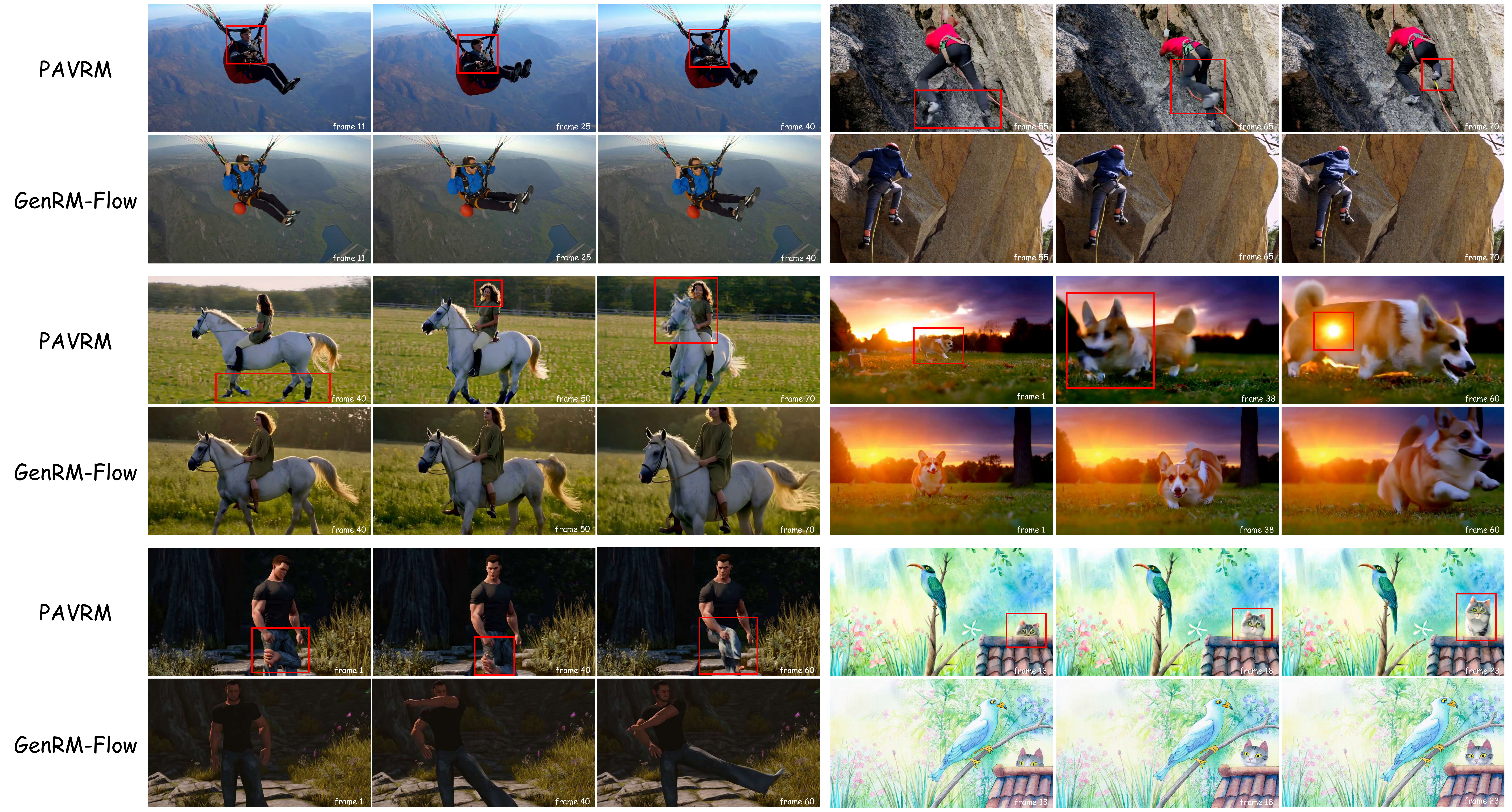}
\caption{Qualitative comparison between PAVRM and \textbf{GenRM-Flow} as the per-step reward signal in FlowAWR with a 1.3B backbone. Both runs use FlowAWR with the same configuration, differing only in the reward signal. Red boxes mark localized structural deformations, spatial inconsistencies, and temporal artifacts produced by PAVRM. Across scenes, our reward yields visibly fewer geometric defects and better spatial and temporal continuity.}
\label{fig:qualitative}
\end{figure*}

\paragraph{Generation Quality with FlowAWR.} To assess whether reward-model accuracy translates into generation gains, we drive FlowAWR fine-tuning with the sequence-level VideoAlign~\cite{liu2025improving} reward and the process-level PAVRM and GenRM-Flow rewards under shared hyperparameters.
\textbf{Quantitative results.} Table~\ref{tab:t2v_comprehensive} reports VBench scores. Our reward yields the best results across all four metrics on both backbone scales, whereas PAVRM yields only marginal gains on the 1.3B model and even slightly lowers VBench Subject Consistency ($97.56 \rightarrow 97.52$), mirroring the prediction-accuracy gap in Table~\ref{tab:rm_comparison}. On the 1.3B backbone, PAVRM even falls behind the sequence-level VideoAlign on three of the four VBench metrics (MS, SC, and HA), whereas GenRM-Flow outperforms both baselines on every metric at both scales.
\textbf{Qualitative analysis.} Figure~\ref{fig:combined_reward} (middle, right) tracks HPSv3 during fine-tuning: ours stays consistently higher, while PAVRM training is lower and more volatile. Qualitatively (Figure~\ref{fig:qualitative}), PAVRM outputs exhibit localized defects, including limb distortion during paragliding and implausible object interactions, while ours produce anatomically and spatially coherent videos.

\begin{table*}[t]
\centering
\caption{Comprehensive quantitative comparison of reward-guided generation on our inner test set and the out-of-domain VBench benchmark. \textbf{Bold} denotes the highest metric value per backbone.}
\label{tab:t2v_comprehensive}
\small
\setlength{\tabcolsep}{5.3pt}
\begin{tabular}{ll cccc cccc}
\toprule
\multirow{2}{*}{\textbf{Base Model}} & \multirow{2}{*}{\textbf{Method}} & \multicolumn{4}{c}{\textbf{Inner Test Set}} & \multicolumn{4}{c}{\textbf{VBench}~\cite{huang2023vbench}} \\
\cmidrule(lr){3-6} \cmidrule(lr){7-10}
& & \textbf{MS} $\uparrow$ & \textbf{DD}$\uparrow$ & \textbf{SC}$\uparrow$ & \textbf{HA}$\uparrow$ & \textbf{MS}$\uparrow$ & \textbf{DD}$\uparrow$ & \textbf{SC}$\uparrow$ & \textbf{HA}$\uparrow$ \\
\midrule
\multirow{4}{*}{Wan2.1-T2V-1.3B} & Pretrain & 98.31 & 59.07 & 97.02 & 88.15 & 98.52 & 65.19 & 97.56 & 94.00 \\
& VideoAlign~\cite{liu2025improving} & 98.58 & 60.68 & 97.22 & 89.96 & 98.73 & 67.52 & 97.64 & 95.32 \\
& PAVRM~\cite{mi2025video} & 98.56 & 63.59 & 97.18 & 89.73 & 98.71 & 67.83 & 97.52 & 94.68 \\
& \textbf{GenRM-Flow} & \textbf{98.88} & \textbf{72.43} & \textbf{98.28} & \textbf{93.16} & \textbf{98.95} & \textbf{73.46} & \textbf{97.70} & \textbf{96.25} \\
\midrule
\multirow{4}{*}{Wan2.1-T2V-14B} & Pretrain & 98.12 & 61.18 & 97.18 & 89.32 & 98.30 & 65.46 & 97.52 & 95.40 \\
& VideoAlign~\cite{liu2025improving} & 98.26 & 65.32 & 97.45 & 90.58 & 98.51 & 70.36 & 97.60 & 95.62 \\
& PAVRM~\cite{mi2025video} & 98.34 & 68.59 & 97.82 & 91.45 & 98.41 & 71.82 & 97.68 & 96.18 \\
& \textbf{GenRM-Flow} & \textbf{98.76} & \textbf{77.48} & \textbf{98.45} & \textbf{95.08} & \textbf{99.03} & \textbf{74.87} & \textbf{97.73} & \textbf{97.73} \\
\bottomrule
\end{tabular}
\end{table*}

\subsection{Ablation Studies}
\label{subsec:ablation}

\paragraph{Intra-group Reward Normalization.}
We study how the raw process reward $r_t$ (Eq.~\eqref{eq:process_reward}) should be normalized within each group during FlowAWR training, comparing three schemes on Wan2.1-T2V-1.3B: the raw reward without group normalization, Z-score normalization, and Softmax (Eq.~\eqref{eq:softmax_reward}). As shown in Table~\ref{tab:ablation_normalization}, training with the raw reward collapses within 250 steps, which we attribute to the large scale variation of $r_t$ across prompts and timesteps in the absence of group normalization. Both Z-score and Softmax remove the prompt-dependent offset and stabilize training. Softmax further outperforms Z-score on all metrics (e.g., raising VBench Dynamic Degree from $69.85$ to $73.46$), which we attribute to its bounded, nonlinear emphasis on the top-ranked trajectories in each group.

\begin{table*}[!htb]
\centering
\caption{Ablation on intra-group normalization mechanisms during FlowAWR training (Wan2.1-T2V-1.3B, evaluated on VBench). ``Policy Collapse'' indicates training collapse. \textbf{Bold} denotes the best results among all.}
\label{tab:ablation_normalization}
\small
\setlength{\tabcolsep}{9pt}
\begin{tabular}{l l cccc}
\toprule
\textbf{Normalization Mechanism} & \textbf{Training Status} & \textbf{MS}$\uparrow$ & \textbf{DD}$\uparrow$ & \textbf{SC}$\uparrow$ & \textbf{HA}$\uparrow$ \\
\midrule
Raw $r_t$ (no normalization) & Policy Collapse ($\sim$250 steps) & N/A & N/A & N/A & N/A \\
Z-score Normalization & Converged & 98.74 & 69.85 & 97.61 & 95.61 \\
\textbf{Softmax (Ours)} & Converged & \textbf{98.95} & \textbf{73.46} & \textbf{97.70} & \textbf{96.25} \\
\bottomrule
\end{tabular}
\end{table*}

%% file: sec/Conclusion.tex
\section{Conclusion}
\label{sec:conclusion}

We have introduced \textbf{GenRM-Flow}, a reward modeling framework in which a video generator, trained on preference data through standard preference optimization, serves as a process-aware reward model in flow matching. The reward signal is sourced from the generator's own generative capacity rather than from an external scorer or an auxiliary head, and is computed from the velocity predictions of the trained and reference models at any timestep along the denoising trajectory. On held-out preference pairs, it reaches $88.96\%$ and $86.79\%$ accuracy on the 14B and 1.3B backbones, surpassing the feature-extractor baseline by $6.32$ and $10.73$ points, respectively. As a per-step process reward source in FlowAWR, the resulting reward model drives measurable downstream generation improvement.

\paragraph{Limitations and future work.}
GenRM-Flow evaluates the process reward on states obtained by forward-noising the generated end-state, rather than on the intermediate states actually visited during sampling. The reward therefore reflects how well the final generation is supported at each noise level, rather than the quality of the specific denoising trajectory that produced it. Extending the reward to the on-trajectory intermediate states is a natural next step.